\documentclass[letterpaper, 10 pt, conference]{ieeeconf}  
\AtBeginDocument{\let\autocite\cite}
\let\tightlist\relax

\IEEEoverridecommandlockouts                              

\usepackage{balance}
\usepackage{cite}
\usepackage{graphicx} 
\usepackage{supertabular}
\usepackage{flushend}
\usepackage{tikz}
\newcommand\copyrighttext{%
  \footnotesize \textcopyright 2012 IEEE. Personal use of this material is permitted.
  Permission from IEEE must be obtained for all other uses, in any current or future
  media, including reprinting/republishing this material for advertising or promotional
  purposes, creating new collective works, for resale or redistribution to servers or
  lists, or reuse of any copyrighted component of this work in other works.}
\newcommand\copyrightnotice{%
\begin{tikzpicture}[remember picture,overlay]
\node[anchor=south,yshift=10pt] at (current page.south) {\fbox{\parbox{\dimexpr\textwidth-\fboxsep-\fboxrule\relax}{\copyrighttext}}};
\end{tikzpicture}%
}

\begin{document}

\IEEEoverridecommandlockouts
\overrideIEEEmargins

\title{\LARGE \bf
On the causality between affective impact \\ and coordinated human-robot reactions
}

\author{Morten Roed Frederiksen$^{1}$ & Kasper Stoy$^{2}$
  \thanks{$^{1}$Morten Roed Frederiksen {\tt\small mrof@itu.dk} and $^{2}$Kasper Stoy {\tt\small ksty@itu.dk} are affiliated with the REAL lab at the Computer science department of
        The IT-University of Copenhagen, Rued Langgaards vej 7, 2300 Copenhagen S
        }
}

\maketitle

\begin{abstract}
In an effort to improve how robots function in social contexts, this paper investigates if a robot that actively shares a reaction to an event with a human alters how the human perceives the robot's affective impact. To verify this, we created two different test setups. One to highlight and isolate the reaction element of affective robot expressions, and one to investigate the effects of applying specific timing delays to a robot reacting to a physical encounter with a human. The first test was conducted with two different groups (n=84) of human observers, a test group and a control group both interacting with the robot. The second test was performed with 110 participants using increasingly longer reaction delays for the robot with every ten participants.
The results show a statistically significant change (p$<$.05) in perceived affective impact for the robots when they react to an event shared with a human observer rather than reacting at random. The result also shows for shared physical interaction, the near-human reaction times from the robot are most appropriate for the scenario.
The paper concludes that a delay time around 200ms may render the biggest impact on human observers for small-sized non-humanoid robots. It further concludes that a slightly shorter reaction time around 100ms is most effective when the goal is to make the human observers feel they made the biggest impact on the robot.
\end{abstract}
\copyrightnotice
\section{Introduction}

Creating robots that can understand and express emotions is a
many-faceted problem. One of the many challenges lies in designing a
relatable robotic behavior with which people will want to interact. If
we disregard digital communication channels, robots convey information
through simple means of expression that includes: Sound, appearance,
movements, and gestures \autocite{frederiksenstoy2019}. These means can
improve how well the intentions of the robot are understood, and
correctly timing when to use them can further improve the interaction
and can influence how the robot is perceived
\autocite{Hoffman2014TimingIH}. A lot of research has focused on the
expressive abilities of robots and have so far accomplished making
people recognize robotic expressions of emotions using morphological
attributes
\autocite{Boccanfuso2015AutonomouslyDI,Stiehl2006TheHA,Singh2013ADT},
facial features
\autocite{Bennett2013PerceptionsOA,Benson2016ModelingAV,Lisetti2004ASI,Park2006NeurocognitiveAS,Breazeal2003EmotionAS,Gockley2006ModelingAI,Lee2007NaturalEE,Yim2009DesigningC},
movement
\autocite{Yang2013DevelopmentOE,Yoshioka2015InferringAS,Saerbeck2010PerceptionOA},
orientation \autocite{Bethel2009PreliminaryRH,Lin2013VersatileHR} sound
\autocite{Lisetti2004ASI,Gonsior2012AnEA,Addo2014ApplyingAF,Winkle2017InvestigatingTR,beckerasano2009LaughterIS,Lui2017AnAM,Zhang2017CommunicationAI},
and gestures
\autocite{Yim2009DesigningC,Cohen2011ChildsRO,Chen2011TouchedBA,Singh2013ADT,Xu2015EffectsOA,Xu2013TheRI,Addo2014ApplyingAF,Johal2015CompanionRB,Rincon2018ExpressiveSW,Park2007AnEE,Bethel2008SurveyON}.
When it comes to expressing affective information and standard emotions,
many projects focus on how to maximize comprehension. Relatively few
projects in comparison focus on the impact of delaying when the
expressive features of the robot are used, and how the causality between
participant and robot reactions can affect how the affective information
is conveyed. Michael 2010 proposes how perceived shared emotions can
facilitate coordination between interacting humans without either of
them possessing previous knowledge of intentions \autocite{Michael2010}.
This paper focuses on whether this effect is equally present in
human-robot-interactions and investigates the following:

\begin{itemize}
\tightlist
\item
  If there is a causality between reaction coordination and perceived
  affective impact on a robot. In other words: When humans and robots
  react to the same event, will the humans perceive the robots'
  reactions as stronger?
\item
  Whether delaying the reactions of a robot in a physical conflict
  interaction can strengthen its perceived affective impact.
\end{itemize}

Gaining knowledge on these aspects of expression abilities is something
all areas of robotics can benefit from. The investigation may provide an
answer to when and how robots should behave in order to strengthen the
affective impact of an interaction. This could be beneficial in
situations where robots are required to convey vital information as
efficiently as possible. E.g. Socially assistive robotics and rescue
robots that operate in demanding working environments may be vastly
improved if we, by altering how and when they use their communicative
features, can make them communicate better in a critical situation.

\begin{figure}[!htb]
\minipage{0.24\textwidth}
  \includegraphics[width=\linewidth]{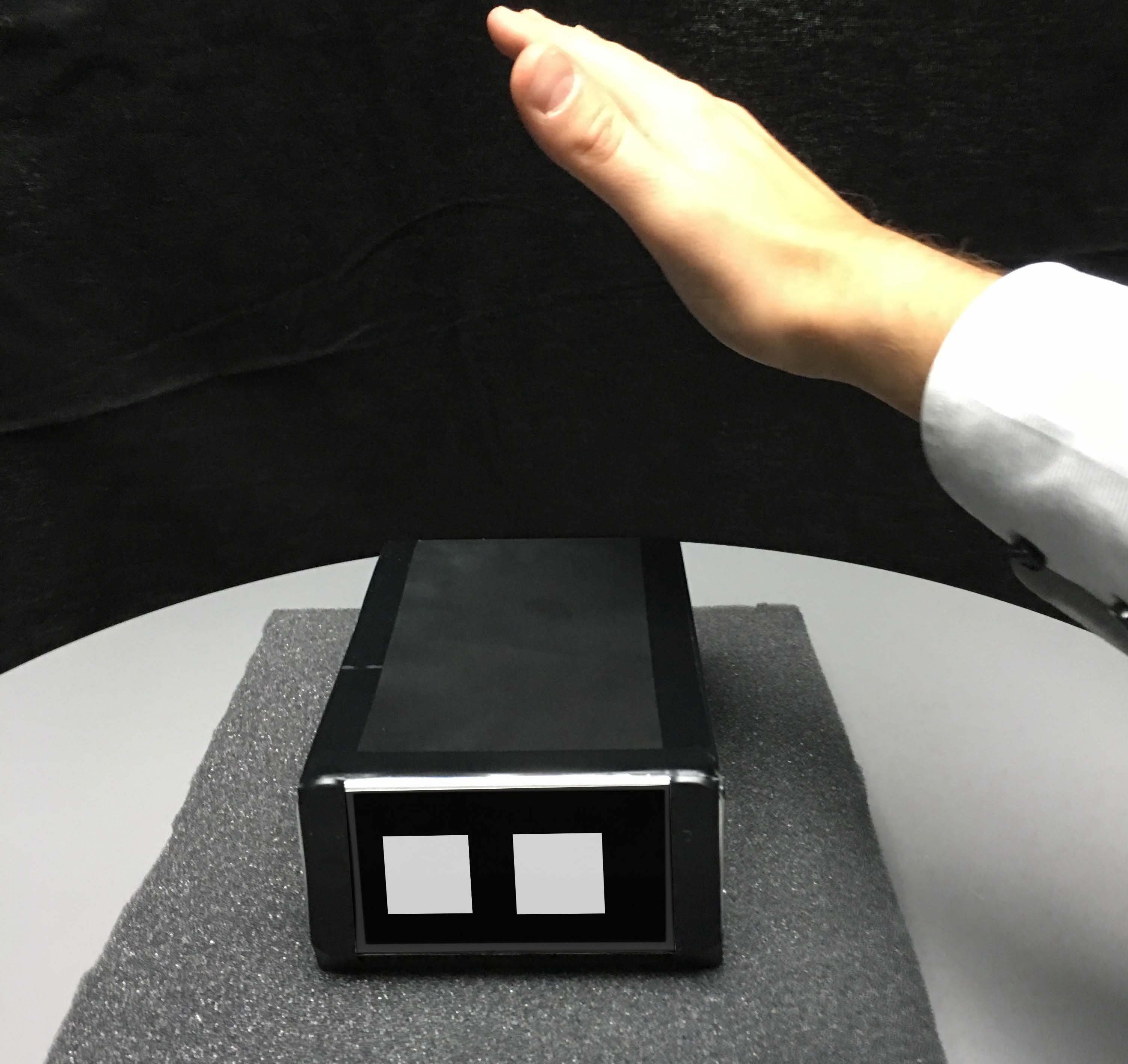}
\endminipage\hfill
\minipage{0.24\textwidth}
  \includegraphics[width=\linewidth]{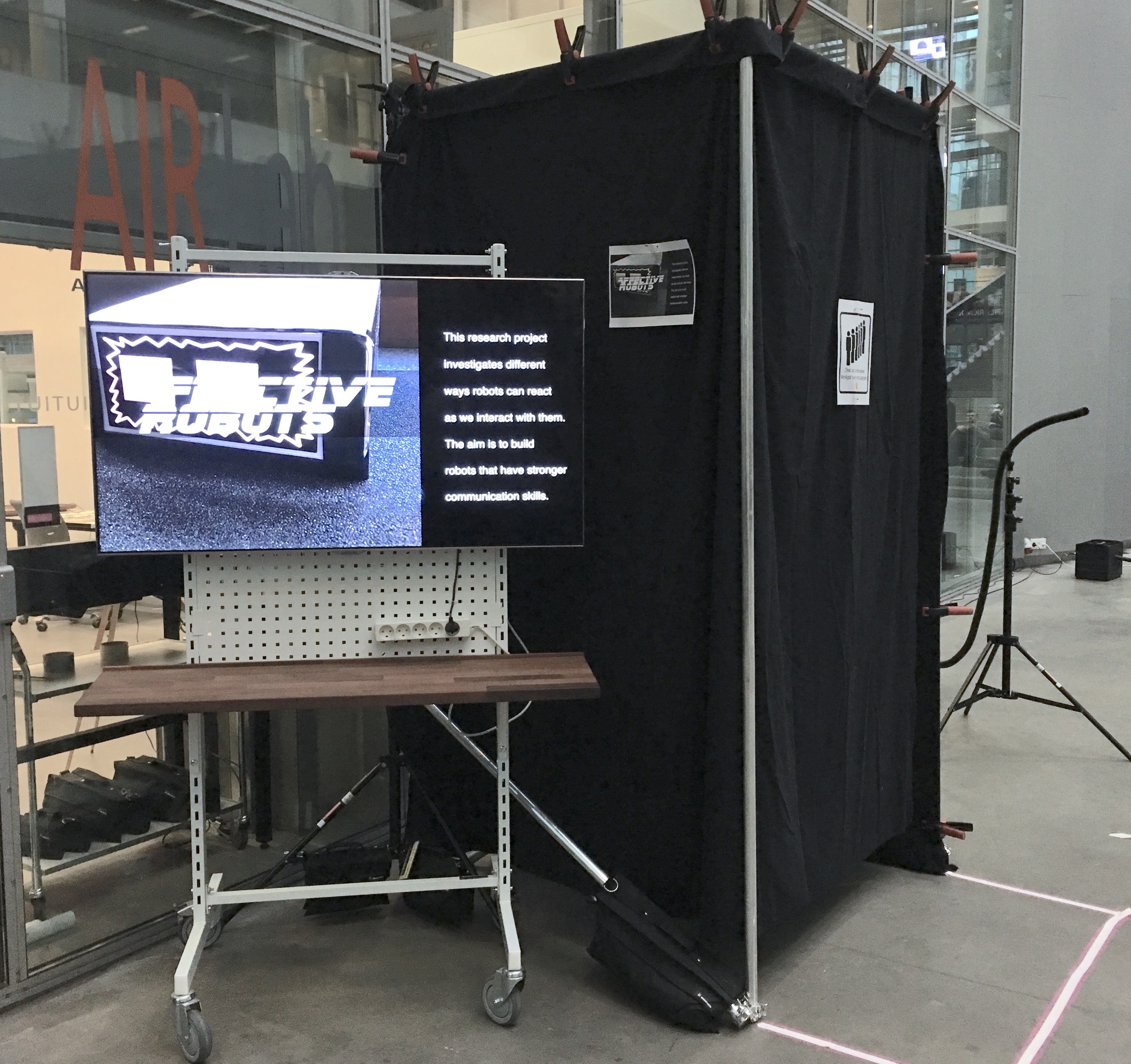}
\endminipage
\caption{Left: The "Affecta" robot. The robot was fastened to a soft foam pad to hinder it from moving as people interacted with it. Right: The test setup included an isolated room to let the participants interact with the robot undisturbed.}
\label{affecta}
\end{figure}

Through each human-robot interaction, the timing dictates who initiates
actions throughout the encounter. E.g. a swift reacting robot could make
a human recipient hold back in the interaction or a robot that delays
answering could make a human counterpart take charge of the situation.
Among other aspects of communication, the timing encompasses both
estimating when to perform movements (for robots to safely cooperate
with humans) and controlling the flow of dialogue between humans and
robots \autocite{Unhelkar2018HumanAwareRA}, \autocite{Lucignano2013ADS}.

When robots react to something, the reaction highlights the connection
between robot behavior and the context event, and it establishes the
direction for the current communication. E.g. for a robot that is
designed to portray being afraid of a dog in the vicinity, there is a
timed frame of opportunity after the dog initiates an action where the
robot can react. Any reactions applied in connection with the dog's
actions will be perceived as connected to that event or that agent in
the scenario. The robot's reaction will be interpreted in light of the
event and if a human experiences the same event, the shared experience
may be used to establish a connection between the human and the robot.
The reaction time and response in the situation is influenced by the
complexity and familiarity of the event information as outlined in Hyman
1953 \autocite{Hyman1953StimulusIA}. Besides the complexity of event
information to which the robot reacts, the hypothesis is that the
following two things (among other factors) can influence how the
expression of a robot's reaction is perceived:

\begin{itemize}
\tightlist
\item
  The time delay with which the robot reacts
\item
  If the reaction is shared with someone.
\end{itemize}

To investigate this, we used two experiments. The first test was a
standard A-B test aimed at isolating the effects of coordinating
human-robot reactions to a context event, while the second test focused
on how reaction delays affected the shared experience in a physical
interaction. Our findings show a causality between human-robot reaction
coordination and the perceived arousal level of the test robots, with a
statistically significant (p\(<\).05) difference between the main group
and the control group. The results further indicate that the reaction
times of the robots in physical interactions influence the affective
state of the humans interacting with it. We argue that near human-like
reaction reflexes overall have the biggest affective impact on the test
participants, while a slightly lesser delay time (
\textasciitilde{}100ms faster) should be used when the aim is for the
test participants to feel they made a big impression on the robot. The
results also indicate that the perceived affective impact of the robot
is strengthened slightly by delaying the reaction.

The presented findings are novel in that they present a new context for
using shared experiences to gain emotional coordination in human-robot
interaction scenarios. The new approaches are based on using
non-humanoid robots and by placing participants on the opposite side of
the robot in a high-intensity conflict situation. The results introduce
many opportunities for further research on the topic. As a whole, they
suggest investigating to what extend shared reactions could strengthen
the affective expression abilities of rescue robots and improve the
reception of critical messages in high-intensity contexts.

\begin{figure*}[t!]
  \includegraphics[width=\textwidth]{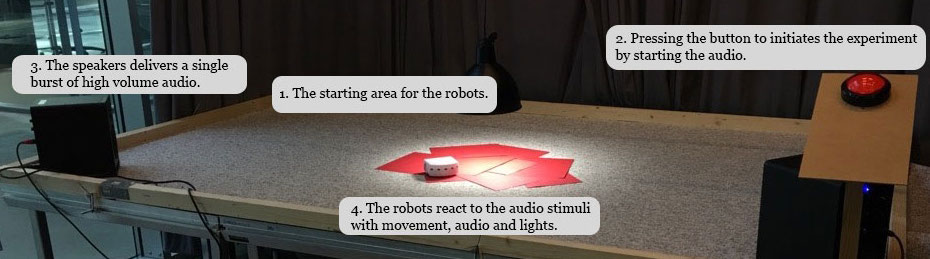}
  \caption{The test setup for the initial experiment. The red center of the arena marks the starting position for the robots. The red button on the right side initiates the experiment in the first test. The same button was removed for the control test of the experiment in which the robots reacted with random intervals.}
  \label{test_surface}
\end{figure*}

\section{Other approaches}

The timing aspects of cooperative interaction was investigated by Pan et
al.~2019 \autocite{pan2019} by in- and decreasing the reaction times of
a robot that was handed an object. The study, which used a humanoid
torso robot with a head and arms, found that the people preferred
reaction time equal to normal human reaction time when interacting with
the robot. Their test scenario was different than the scenario
investigated in this paper, as it contained a low-intensity interaction,
a humanoid robot, and a cooperative task to accomplish in the tests,
whereas this project focuses on non-humanoid robots in a high-intensity
scenario and a test task that emphasizes the conflict between the
interacting human and robot participants.

Previous robot projects have investigated increasing the understanding
of affective communication in their research. Brazeal et al.~2003
employed an emotional subsystem for the robot Leonardo and controlled
realistic employment of several affective means of expression making it
easier to understand \autocite{Breazeal2003EmotionAS}. Gunes et al.~2011
used a LEGO-based custom robot to convey the emotional intentions of
classical music. The robot employed several affective means of
expression including movement and onboard gestures to communicate the
affective status \autocite{gunes2011}. The timing aspects were the focus
of Huber et al.~2008 in which they investigated different ways of
letting robots hand over objects to humans. Successfully handing over
the objects requires both parties of the interaction to agree on a
common timing for the involved movements. The study found that the less
jerky the movement was, the safer they felt around the robot.
\autocite{Huber2008HumanrobotII}.

Bing and Michael 2012 investigated how sharing a stressful experience
with a humanoid robot can potentially help humans overcome the uncanny
valley effect \autocite{bing2012,Mori2017TheUV}. The 2012 paper found
that their test participants preferred familiar humanoids with whom they
had shared a stressful experience with rather than familiar robots that
they had shared a pleasant experience with. This paper aims to extend
the results found in that paper on two different levels. It investigates
whether the results are similar for a non-humanoid robot that bears no
resemblance to a person, and it attempts to discover whether the result
is isolated to people that are on opposite sides of a conflict- and
stressful situation. This paper emphasizes how humans perceive robot
specific nonverbal behavior which is also the focus in Putten et
al.~2018 \autocite{Ptten2018TheEO}. In this paper, the robot-like
specific behavior is found less effective than using human-like familiar
behaviors to convey affective information. Both Bing \& Michael 2012 and
Pûtten et al.~2018 indicates the strengths of using human-inspired
behaviors and morphology in their studies which makes a good contrast to
the experiments performed in this paper using non-humanoids and strictly
robot-specific behaviors.

\section{Method}

The first test aims at investigating changes to the general composition
of emotions, while the second test expands the investigation into a
physical and confrontational scenario to see how that influences the
perceived intentions of a robot. The second test also focuses on the
immediate delay between the context event and the subsequent robot
reaction to see how delaying the robot's reaction influences how the
robot was perceived. As stated in Bing et al.~2012, a shared stressful
event works stronger using humanoid robots, which is why a conflicting
scenario with a non-humanoid robot was interesting for the second test
in this paper \autocite{bing2012}.

\subsection{Using standard descriptors}

In affective robotics research it is often the Pleasure, Arousal and
Dominance (or PAD) scale that is used to describe emotional states
\autocite{russel1974},\autocite{Russell2003CoreAA}, while temporal
aspects can be classified in the Traits, Attitudes, Moods and Emotions
(TAME) architecture \autocite{Moshkina2011TAMETA}.

We quantify the affective impact by measuring the changes to the robot's
perceived current emotion in PAD space. We measure differences between
the two test groups on how the robot's affective state is perceived. If
the test participants find it more or less pleasant, aroused, or
dominant. E.g. if a person is angry during an interaction with a robot,
and the robot emits a soothing sound to make the person change to a
happier state, the angry emotion could move along the `arousal' axis
towards less aroused - which would be considered an affective change to
the current affective state. This is what we use as a quantitative
measure for the effects of coordinated reactions in the initial tests.

The tests followed a standard A-B pattern with two individual groups of
test participants where one of them acted as a control group. The two
groups would encounter the same scenarios, but the control group of
participants would not experience coordinated reactions with the robots
as they would react at random and out of phase with the participants.
The test setup is depicted in Figure \ref{test_surface}.

\subsection{Moving to a physical interaction}

Building upon the outcome of the first tests, the second test focused on
how the shared reaction was perceived when the interaction context was
changed to a physical and conflicting encounter with closer proximity
between the participants and the robot. In this test, we asked the
participants to physically strike the robot as much as they wanted and
observe the reaction. We departed from using the standard PAD descriptor
as we were not focusing on the composition of the affective impact, but
rather on investigating where the interaction was perceived as making
the largest impact - on the robot itself or the test participants. We
also wanted to see how the delay time influenced the perceived size of
the affective reaction and introduced delays between the physical
interaction and the robot's reaction to highlight the connection between
them. As the robot reacted in this context, the swiftness of the
reaction made it more similar to a reflex than a prepared response. This
approach was chosen as it matched the conflicting scenario. The sharing
in the second test was solely the interaction, and we attempted to
investigate how placing the participants and the robot on opposite sides
of a conflict situation influenced the human-robot relationship.

\section{Experimental setup}

In the first test, there were two groups with 42 people observing the
robots in each of them. The overall gender distribution was 39 females
and 45 males in ages from 10 to 50+. The majority of the participants
were between 20 and 30 (71\%) years old, and most of the participants
either worked - or studied at The IT-University of Copenhagen (82.5\%).

In the second test there were 110 participants distributed in 7 groups.
The gender distribution here was 56\% male and 44\% female and the
largest age group was 20-30 years old (33\%) followed by people between
10-20 (20.2\%). The initial test used a ``Thymio 2'' robot while the
second test used and altered a custom-built ``Affecta'' robot designed
to convey affective information.

\subsection{The first test: impact of reaction}

The setup of this test was comprised of three ``Thymio 2'' robots and a
designated arena for the robots to move on. The arena was constructed
from stage parts, forming a 220cm times 300cm surface, with floor
carpets on top to create a smooth surface to easily maneuver on for the
low-clearance Thymio 2 robots. The edges of the designated test arena
were padded with a small wooden edge to prevent the robots from falling
to the ground. The edges were fastened just high enough to trigger the
proximity sensors positioned at the front, side, and back of the robots.
The first test contained two experimental phases with different groups
participating in each experiment. The tests were initiated in isolation
from each other and followed this test outline:

\begin{figure}[h!]
  \includegraphics[width=8.8cm]{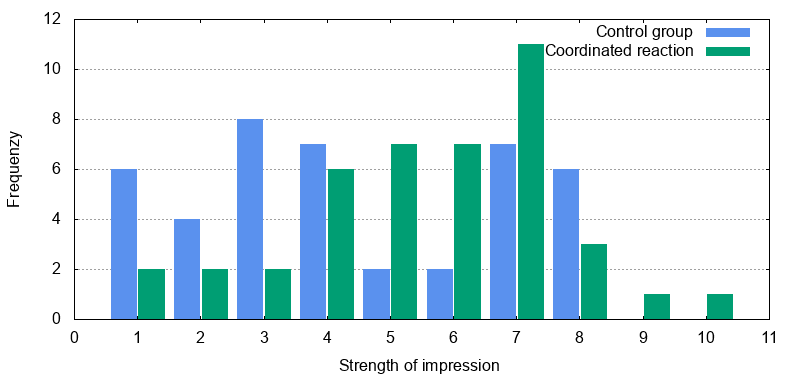}
  \caption{The diagram shows the perceived arousal level of the robots. The blue-colored values are from the control test with uncoordinated human-robot reactions while the red-colored values are from the test group coordinated reaction between the participants and the robots.}
  \label{agitated}
\end{figure}

\emph{Test steps:} \newline 1. The robots were initially placed at the
center of the arena. See Figure \ref{test_surface} for the initial
position of the robots. 2. The participants were asked to start the
experiment by pushing a button. 3. A high volume sound of an explosion
was played as the button was pushed and the robots (and participants)
reacted to the sound displaying fear. The robots used the following
expression modalities: Sound, movement, and colored lights to convey the
fear behavior. 4. The robots moved from the start area with maximum
speed while displaying lights, and playing alerting audio signals in an
attempt to show fearful behavior. 5. After 2 seconds of employing audio
and lights, the robots continued to move but the audio and lights were
turned off. This was done to enforce the connection between the reaction
and the event that initiated it. 5. The robots moved randomly around on
the surface while using front and back sensors to avoid the perimeter.
6. Once the robots encountered the center `resting' area again, they
stopped and waited until reacting again (start over from point 2).

The control group would go through the same steps. However, the robots
would not react in coordination with the sound but at random intervals.
After each experiment, the test participants were asked how
\emph{aroused} they perceived the robots were, how \emph{pleasant} they
perceived the robots found the experiment, and how \emph{dominant} they
perceived the current emotion for the robots was on a scale from 1 to 10
(1 meaning: not at all and 10 meaning: maximum possible). The
participants were additionally asked to state their gender, and age.

\subsection{The second test: the impact of specific timing}

The second test used a custom-built robot as depicted in the left image
of Figure \ref{affecta}. The robot was is a small non-humanoid
box-shaped robot that was designed to have implementations for a large
variety of expression modalities, making it a great fit for this
project. This specific robot design was 3d-printable, and suited the
test setup. For the robot to remain stable for the physical interaction,
only the top part of the robot was used and the bottom drive wheels not
added. The robot consisted of two separate software architectures - a
ROS based part to control the physical movement and gestures of the
robot and a mobile application with access to all available sensors on a
mobile smartphone. For this test, the mobile IOS based platform was
expanded with a module for detecting physical movement using the onboard
accelerometer. When the user would hit the robot the accelerometer
sensor was triggered which informs the main robot controller to display
a reaction using the mobile phone screen and audio capabilities of the
robot (also supplied by the phone). The reaction consisted of a loud
alert noise and jagged lines flashing at the edge of the screen. The
second test was set up in a specially constructed and isolated test
booth. The booth, which can be seen in the right image of Figure
\ref{affecta}, contained a table with the robot at a raised position to
facilitate a close proximity interaction, and it contained a poster with
instructions for the test participants to strike the robot. One at a
time we asked them to enter the test booth and hit the robot as much as
they liked. They would interact with the robot by hitting it and observe
how the robot reacted. When the test participants were finished with the
physical interaction, they would step outside of the test booth and we
proceeded by asking the following questions:

\begin{itemize}
\tightlist
\item
  How big an impact did your actions make on the robot?
\item
  How big an impact did the robot's reaction make on you?
\item
  How appropriate would you rate the robot's actions as being in light
  of how you interacted with it?
\end{itemize}

The participants were also asked to state their age group and their
gender. The test was completed with 110 test participants. With each
group of ten participants, the reaction delay of the robot's reactions
was doubled starting from an initial reaction delay of 50ms ending at a
reaction delay of 3200ms.

\section{Results}

The first test isolated the effects of coordinating human-robot
reactions to a context event, while the second test used increasingly
longer reaction delays to investigate how that affected the perceived
affective impact of a human-robot physical interaction.

The results show three important findings:

\begin{itemize}
\tightlist
\item
  There is a causality between coordinating the reactions of humans and
  robots and the perceived arousal level of the test robots.
\item
  The reaction times of the robots in physical interactions influence
  the affective state of the humans interacting with it and near
  human-like reaction reflexes (\textasciitilde{}250ms) have the biggest
  affective impact on the test participants
\item
  A slightly lesser delay time ( \textasciitilde{}100ms faster) is
  preferred when the aim is for the test participants to feel they made
  a big impression on the robot.
\end{itemize}

\subsection{The influence of coordinating human-robot reactions}

In the first test, we asked the participants to rate how \emph{aroused}
the robots seemed, and the difference between levels of perceived
arousal was statically significant (Two-tail Wilcoxon signed-rank,
p\(<\).05). This shows a strong connection between experiencing a shared
reaction with the robots and the interpreted level of arousal conveyed
by the robots. The distribution of answers for the question on the
perceived level of arousal can be seen in Figure \ref{agitated}, and the
key figures for the same question can be seen in Table \ref{averages}.

\begin{table}[]
\centering
\begin{supertabular}{|l|c|c|}
\hline
                             & \multicolumn{1}{l|}{No Reaction (avg/dev)} & \multicolumn{1}{l|}{With Reaction} \\ \hline
Agitatedness       & 4.40/2.42                                  & 5.55/2.04                                    \\ \hline
Pleasantness       & 4.83/2.27                                  & 4.71/2.11                                    \\ \hline
dominance & 3.76/2.34                                  & 3.98/2.50                                    \\ \hline
\end{supertabular}
\\
\caption{The averages and standard deviation for the answers for the perceived level of arousal, pleasantness, and level of dominance in the tests where the participants shared a reaction with the robots and the control test in which the robots reacted at random intervals. }
\label{averages}
\end{table}

We also asked the participants to rate the perceived pleasantness of the
experience for the robots. The results for that question showed no
relevant differences between the random group and the reaction group.
The participants agreed that the experience was mildly unpleasant for
the robots in both groups with key figures as seen in Table
\ref{averages}. The last question regarded the perceived level of
dominance for the current emotion, on which the participants rated each
group with near similar scores. This indicated that there was no
connection between the dominance level and sharing a reaction or not.

\subsection{Reaction delays strengthen affective impact}

\begin{figure}[h!]

  \includegraphics[width=8.8cm]{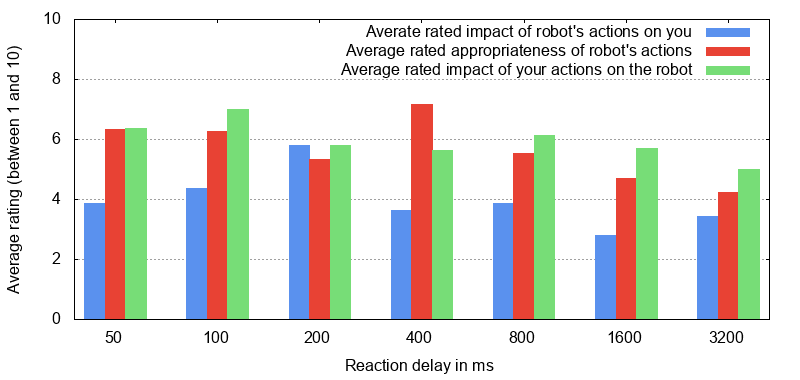}
  \caption{The resulting average ratings in relation to the delay time in milliseconds concerning the rated impact of the robot's actions, the impacts participants made on the robot, and the rated appropriateness of the robot's actions.}
  \label{graph_all}

\end{figure}

In the second test, the results indicate that there was a preferred
reaction delay around 200ms for the question regarding the perceived
impact of the robot's actions on the participants who interacted with
it. The resulting averages for that question can be seen in Figure
\ref{graph_all}. This enforced the results found in by Pan et al.~2019
and extends the finding to also include non-humanoid robots and a
conflicting scenario rather than a cooperative context
\autocite{pan2019}. The results show that the robot made the biggest
affective impact on the participants when it reacted to the physical
interaction with human-like reaction times (which we assume is
approximately 250ms). It is important to state that although our number
of participants is relatively high (n=91), using the arithmetic mean for
smaller individual groupings could make the result more easily affected
by outliers.

We asked the participants to rate how big an impression the test
participants' actions made on the robot, and for that question, the
relative highest rated delay time was 100ms. This and the previous
result indicates the following:

\begin{itemize}
\tightlist
\item
  If the aim is for the robot to make a big impression on the
  participants, it should react with near-human reaction times.
\item
  If the aim is for the test participants to feel they made a big
  impression on the robot, it could benefit from reacting with a
  slightly smaller delay. ( \textasciitilde{}100ms faster).
\end{itemize}

We also asked the participants to rate the appropriateness of the
robot's action in relation to the actions performed by the test
participants. The resulting ratings were near at par with each other
with a reaction time of 100ms rated relatively highest. The resulting
averages for the last two questions can be seen in Figure
\ref{graph_all}.

Grouping the results by the age of the participants shows that most of
the age groups prefer human-like reaction times. The top-rated of the
average reaction time for the affective impact of the robot's behavior
in regards to age group was 200ms. Our initial assumption was that the
results would support a relationship between older age and slower
preferred reaction times. This is however not the case. The 200ms delay
which corresponds to human-like reaction times is preferred even by the
older test participants. The age group from 21 - 30 preferred the
slowest reaction time of 3200ms, but a closer look at the data reveals
that may be explained by a lack of proper age distribution for some
delay times. It is vital to state that the age distribution across every
delay group is not uniformly distributed. Some delay categories has very
few examples for specific age groups. The results indicate that the age
group of 41 - 50 prefers a slower-than-human robot with a preferred
reaction time of 400ms and presents an opportunity for further research
projects to focus more on each age group and the preferred reaction
times.

\section{From measures to meaning}

The boundaries of each discrete state in models such as the PAD space
are fuzzy, and a single 3d coordinate can rarely convey the rich sources
of information that affective data is \autocite{gunes2011}. Because
emotions are given significance by the words that express them, they
differ between languages. In some cases with specific languages, certain
emotions are not present or mean something different
\autocite{Mesquita1992CulturalVI}. When the interpretation and
comprehension of the affective states are culturally dependent the
problem is that the interpretation of them change with each cultural
context and group of human observers \autocite{Gunes2010AutomaticDA}.
This paper acknowledges that it is difficult to create a test setup that
provides clear answers, but attempts to work around it by using many
participants. Our test setup had the two following drawbacks regarding
the age of the participants:

\begin{enumerate}
\def\labelenumi{\arabic{enumi}.}
\tightlist
\item
  The test was designed to measure the effect of the delay times. This
  meant that the age groups were not uniformly distributed within each
  tested delay times and that some delay times had one or more age
  groups that were not represented.
\item
  As our delay time was doubled each time, it left out too many details
  of the interesting area between 200 and 400ms. It may be that the
  effect we were attempting to verify was smaller than anticipated and
  that we instead needed a test that expanded the knowledge on that
  specific delay interval.
\end{enumerate}

The results of the first test indicate that there is a causality between
the level of perceived arousal and the coordination of human-robot
reactions to the context event. The robots were perceived as being more
aroused when their reactions were coordinated with human observers. The
results show that considering the timing aspects of conveying affective
information and sharing a reaction with a human observer can be
beneficial in those scenarios where the aim is to convey highly aroused
affective states.

That the overall voted most suitable reaction delay time for the
reaction to the physical interaction of the second test is 200ms, might
for some scenarios be considered a positive result. Such a delay leaves
a wide timeframe even for low hardware-driven robots to analyze the
input and consider the proper reaction to a given situation. The
physical properties in the second test also seemed to affect how the
participants interpreted the overall pleasantness of the interaction.
Some participants stated they felt bad about hitting the robot and did
not want to interact with it because it seemed as if they punished the
robot for no reason. The average ratings on appropriateness in relation
to reaction time can be seen in Figure \ref{graph_all}.

The resulting ratings for the different delay times fortify what Pan et
al.~2019 found with humans and robots interacting in a cooperative
setting \autocite{pan2019}. Our common intuition would say that the Pan
et al.~test participants preferred a human-like response time because
they used a humanoid robot and a human-to-human inspired context with a
cooperative task. However, if we interpret the highest-rated suitable
behavior as the preferred behavior, our result shows that these findings
can be extended to non-humanoid robots as well. They also show that the
same reaction time was found most suitable in high-intensity scenarios -
in which people physically interact with the robot.

When we asked the participants to rate the emotional impact of hitting
the robot, the highest average rating was given when the robot reacted
with a delay time of 100ms followed by the second-highest ratings for
50ms. This could indicate that there is a measurable difference between
how the participants wanted the robot to react in the different
scenarios. When the aim is to convey to the participants that their
actions had a large impact, the reaction time should be shorter than
human reaction times (\textless{}250ms). When the aim is for the robot
to make a large emotional impact on the participants, the robot should
react similarly to humans (\textasciitilde{}250ms). It makes sense to
consider to what extent the results are applicable in other contexts.
The tested scenario portrayed a social context, and it may be that the
highest-rated reaction speeds in this experiment would be found suitable
for other social situations as well. However, the results do not per se
extend to other robot types and or other domains. E.g. we don't
necessarily prefer a manufacturing robot at a factory to work at the
same speeds as humans.

Regarding the results grouped by age, we argue that the presented
findings introduce many opportunities for further research on the topic.
As one, we suggest investigating more specifically to what extend age
influences the chosen most suitable reaction times in a finer interval
between 200 and 400 ms - to see if age specific reaction times could
strengthen the reception of affective information even further.

\section{Conclusion}

The paper has investigated the causality between coordinating
human-robot reactions and the perceived affective impact on robots. It
has shown that we can use coordinated reactions to strengthen the way
robots convey affective information. The emphasis was to see whether the
perceived level of intensity in the behavior was increased when a robot
was reacting to a context event in coordination with a human, and to
test whether delaying the specific reaction times in physical
interactions influenced how test participants viewed the affective state
of the robot. We carried out two human-robot interaction tests to
highlight these aspects of human-robot interaction.

The result showed that there was a significant difference between how
aroused the human observers rated the robots as being in the first test
when the human-robot reactions were coordinated. The results of the
second test indicated that even for high-intensity scenarios with
non-humanoid robots, the preferred reaction for the robots was similar
to the reaction time of humans. Furthermore, they showed that a faster
reaction time ( \textasciitilde{}100ms faster) was preferred when the
goal was for the test participants to feel as if they made a large
impact on the robot.

The findings indicate that the concept of sharing reactions and using
near-human reaction delays can be strategically used to influence how
the current affective state of a robot is perceived.

\balance
\bibliography{bibliography}
\bibliographystyle{IEEEtran}

\end{document}